\documentclass[preprint,10pt,authoryear]{elsarticle}
\usepackage{amssymb}
\usepackage[fleqn]{amsmath}
\usepackage{enumitem}
\usepackage{float}
\usepackage{multirow}
\usepackage{siunitx}
\usepackage[table,xcdraw]{xcolor}
\usepackage{array}
\usepackage{soul}
\usepackage{xurl}
\soulregister\cite7
\soulregister\ref7 
\soulregister\citep7 
\soulregister\citet7 
\soulregister\pageref7 
\usepackage{setspace}
\usepackage{booktabs}
\usepackage[section]{placeins}
\usepackage{stfloats}
\usepackage{bbding}
\usepackage[utf8]{inputenc}
\usepackage{url}
\usepackage{pifont}
\usepackage{wasysym}
\usepackage{utfsym}
\usepackage{tabularx}
\usepackage{caption}
\usepackage{graphicx}	
\usepackage{lineno}
\usepackage{appendix}
\usepackage{subfigure}
\usepackage{bm}
\usepackage{makecell}
\newcolumntype{L}[1]{>{\raggedright\let\newline\\\arraybackslash\hspace{0pt}}m{#1}}
\newcolumntype{C}[1]{>{\centering\let\newline\\\arraybackslash\hspace{0pt}\vspace{0pt}}m{#1}}
\newcolumntype{R}[1]{>{\raggedleft\let\newline\\\arraybackslash\hspace{0pt}\vspace{0pt}}m{#1}}
\journal{Arxiv}
\usepackage{hyperref}
\hypersetup{hidelinks,
	colorlinks=true,
	citecolor=green,
	allcolors=black,
	pdfstartview=Fit,
	breaklinks=true}
\begin{document}

\begin{sloppypar}
\begin{frontmatter}
\title{Prior-guided Fusion of Multimodal Features for Change Detection in Optical-SAR Images}

\author[pla]{Xuanguang Liu}
\ead{lxg1688@foxmail.com}
\author[pla]{Lei Ding\corref{cor}}
\ead{dinglei14@outlook.com}
\cortext[cor]{Corresponding author: Lei Ding}
\author[pla]{Yujie Li}
\author[pla]{Chenguang Dai}

\author[pla]{Zhenchao Zhang}
\author[fzu]{Mengmeng Li}
\author[pla]{Ziyi Yang}
\author[pla]{Yifan Sun}

\author[pla]{Yongqi Sun}
\author[syu]{Hanyun Wang}
\author[ita]{Lorenzo Bruzzone}

\address[pla]{Institute of Geospatial Information, Information Engineering University, \\Zhengzhou, China}
\address[fzu]{Academy of Digital China (Fujian), Fuzhou
University, Fuzhou, China}
\address[syu]{The School of Electronics and Communication Engineering, Sun Yat-sen University, Shenzhen, China}
\address[ita]{The Department of Information Engineering and Computer Science, University of Trento, Trento, Italy}

\begin{abstract}
Multimodal change detection (MMCD) identifies changed areas in multimodal remote sensing data, demonstrating significant application value in land use monitoring and urban sustainable development. However, literature MMCD approaches exhibit limitations in both cross-modal interaction and exploiting modality-specific characteristics. This leads to insufficient modeling of fine-grained change information, thus hindering the precise detection of semantic changes. To address these problems, we propose STSF-Net, a framework designed for MMCD between optical and SAR images. STSF-Net jointly models modality-specific and spatio-temporal common features to enhance change representations. Specifically, modality-specific features are exploited to capture genuine semantic change signals, while spatio-temporal common features are embedded to suppress pseudo-changes caused by differences in imaging mechanisms. Furthermore, we introduce an optical and SAR feature fusion strategy that adaptively adjusts multimodal feature importance based on semantic priors obtained from visual foundation models. Finally, we introduce the novel Delta-SN6 dataset, the first openly-accessible multiclass MMCD benchmark consisting of very-high-resolution fully polarimetric SAR and optical images. Experimental results on Delta-SN6, BRIGHT, and Wuhan datasets demonstrate that our method outperforms the state-of-the-art by 3.21$\%$, 0.87$\%$, and 1.32$\%$ in mIoU, respectively.
\end{abstract}

\end{frontmatter}


\section{Introduction}
Multimodal change detection (MMCD) identifies Earth's surface changes through the synergistic fusion of heterogeneous remote sensing data obtained from the same geographical region at multiple time points \citep{BROWN2020111356}. Different from homogeneous change detection \citep{JiangLong,ZhenLi}, by integrating complementary information across modalities, MMCD can effectively overcome limitations inherent in single-source data. Among these, combining optical and Synthetic Aperture Radar (SAR) imagery to detect land cover and land use (LCLU) change is a popular MMCD task. SAR provides all-weather, day-night imaging capability, which mitigates issues like cloud occlusion and supports reliable monitoring in time-sensitive scenarios such as disaster response. Moreover, the fusion of spectral characteristics from optical data and dielectric and surface roughness information from SAR enables a more comprehensive characterization of LCLU changes.

Although optical-SAR change detection (OSCD) has many advantages, the distinct imaging mechanisms result in significant differences in radiometric and textural characteristics between modalities, making direct comparison and alignment of cross-modal features challenging \citep{Yang03092025}. The field of OSCD has recently focused on addressing the critical challenge of cross-modal discrepancies \citep{Lv9955391}. Early post-classification change detection methods \citep{Nielsen4060945} independently segmented optical and SAR images. These methods circumvented the need to address cross-modal discrepancies but ignored bi-temporal spatio-temporal dependencies, introducing significant cumulative errors. With the development of deep learning, research paradigms have shifted toward learning-based feature fusion, primarily divided into transformation-based methods and feature alignment-based methods. The former transforms one modality into another, regarding OSCD as a homogeneous change detection problem. The latter, as the mainstream framework, aims to map multimodal data into a shared feature space to achieve cross-modal feature alignment and subsequently learn change information.

However, existing research emphasizes learning domain-invariant common features, which excessively suppress modality-specific characteristics. In fact, modality-specific features carry critical discriminative information for identifying subtle variations and thus can improve the detection of change targets \citep{Zhao11010882}. Therefore, effectively leveraging complementary information across modalities while maintaining semantic consistency is a critical problem that demands a solution. Moreover, literature OSCD methods exhibit significant limitations in capturing the intricate spatio-temporal dependencies and evolutionary patterns inherent in multimodal observation data \citep{Liu11159545}. These studies predominantly focus on exploiting pixel-level spectral-temporal features while neglecting the spatial structure and contextual semantic information of ground objects. We recognize that LCLU changes are not isolated events, but rather follow structured spatial arrangements and distribution rules, exhibiting distinct transition patterns.

An additional challenge for the advancement of OSCD is related to the scarcity of large-scale, high-quality benchmarks. Existing datasets are often limited in scale, lack fine-grained semantic annotations, and seldom include fully polarimetric SAR data paired with high-resolution optical imagery.

To address these limitations, we identify the explicit disentanglement and joint modeling of modality-specific and common features as a promising pathway, as it allows for fuller utilization of both semantic consistency and complementary information across modalities. The recent emergence of Visual Foundation Models (VFMs), trained on massive datasets and rich with transferable semantic priors, provides a new opportunity. Inspired by this, we propose to leverage the semantic priors from VFMs to guide the multimodal fusion process, thereby steering a more effective integration of these complementary feature representations. Our core contributions are summarized as follows:
\begin{itemize}
\item We propose a dual-branch OSCD framework that explicitly disentangles and jointly models modality-specific and spatio-temporal common features. It comprises a Modal-Specific Feature Encoder (MSFE), tailored to the individual modality characteristics, and a Spatio-Temporal Common Feature Modeling (STCFM) module, which extracts consistent representations across modalities.
\item We introduce a semantic prior-guided multimodal feature fusion mechanism that adaptively fuses the disentangled specific and common features. Leveraging change priors from VFMs as dynamic fusion weights, it selectively amplifies discriminative features to enhance change representations while utilizing modality-common features elsewhere to maintain structural consistency and suppress false alarms.
\item We construct and release Delta-SN6, the first publicly available multiclass OSCD benchmark featuring co-registered, VHR (0.5 m) fully polarimetric SAR and optical imagery. It addresses critical gaps in temporal alignment, annotation granularity, and task applicability, providing a high-quality foundation for advancing multimodal change detection research.
\end{itemize}

\section{Related work}
\subsection{Optical-SAR change detection}
DL-based methods have become the dominant paradigm for OSCD \citep{MA2019166}. These methods bridge the domain gap between heterogeneous modalities through domain adaptation and GAN techniques. Image translation-based methods \citep{ZHUANG2025104321, YANG2022108208} convert SAR images into pseudo-optical images via GANs, transforming MMCD into a homogeneous change detection task solvable by mature methods. Feature-level domain adaptation-based methods \citep{LI202114, 9497508} perform alignment directly at the feature level without requiring pixel-level transformation. Through adversarial training, the network learns sensor-agnostic, domain-invariant feature representations. However, these approaches prioritize learning common features at the expense of modality-specific characteristics that provide critical discriminative information for detecting subtle semantic changes.

\subsection{Multimodal feature fusion}
Attention mechanisms have been widely integrated to enhance multimodal feature fusion, with representative approaches including multi-scale attention \citep{10103595, 9632564, rs14143464}, high-frequency attention \citep{ZHENG2022108717, HOU2026116042}, and methods based on deeply supervised attention metrics \citep{9467555, Han2024}. More recently, powerful architectures such as Transformers and state space models have offered robust frameworks for deep feature interaction \citep{Liu2024, LIU202573, TONG2026114928}. However, these methods generally apply a uniform interaction pattern across all spatial locations, failing to explicitly differentiate between high-probability change regions and unchanged regions, which results in inadequate suppression of pseudo-changes and limited sensitivity to real changes.

\subsection{VFMs-based change detection}
VFMs such as SAM and CLIP, trained on ultra-large-scale datasets, possess powerful visual representations and zero-shot generalization capabilities, offering rich semantic priors about objects and scenes. VFM-based OSCD methods can be categorized into three approaches. First, frozen VFMs serve as generic feature extractors that derive multi-scale features from optical and SAR images for subsequent difference analysis and classification \citep{yu2024exploringfoundationmodelsremote}. Second, prompting-based methods leverage the reasoning capabilities of VFMs \citep{DONG202453}, though their performance depends heavily on prompt quality and the generalization of VFMs to the RS domain. Third, fine-tuned VFMs serve as CD network encoders \citep{10443350, 10543161, dong2025peftcdleveragingvisionfoundation}, using parameter-efficient fine-tuning on CD data and modules tailored for temporal difference and multi-scale information processing.

\section{Methodology}
\subsection{Overview of the STSF-Net}
\begin{figure*}[h]
    \centering
    \includegraphics[width=\linewidth]{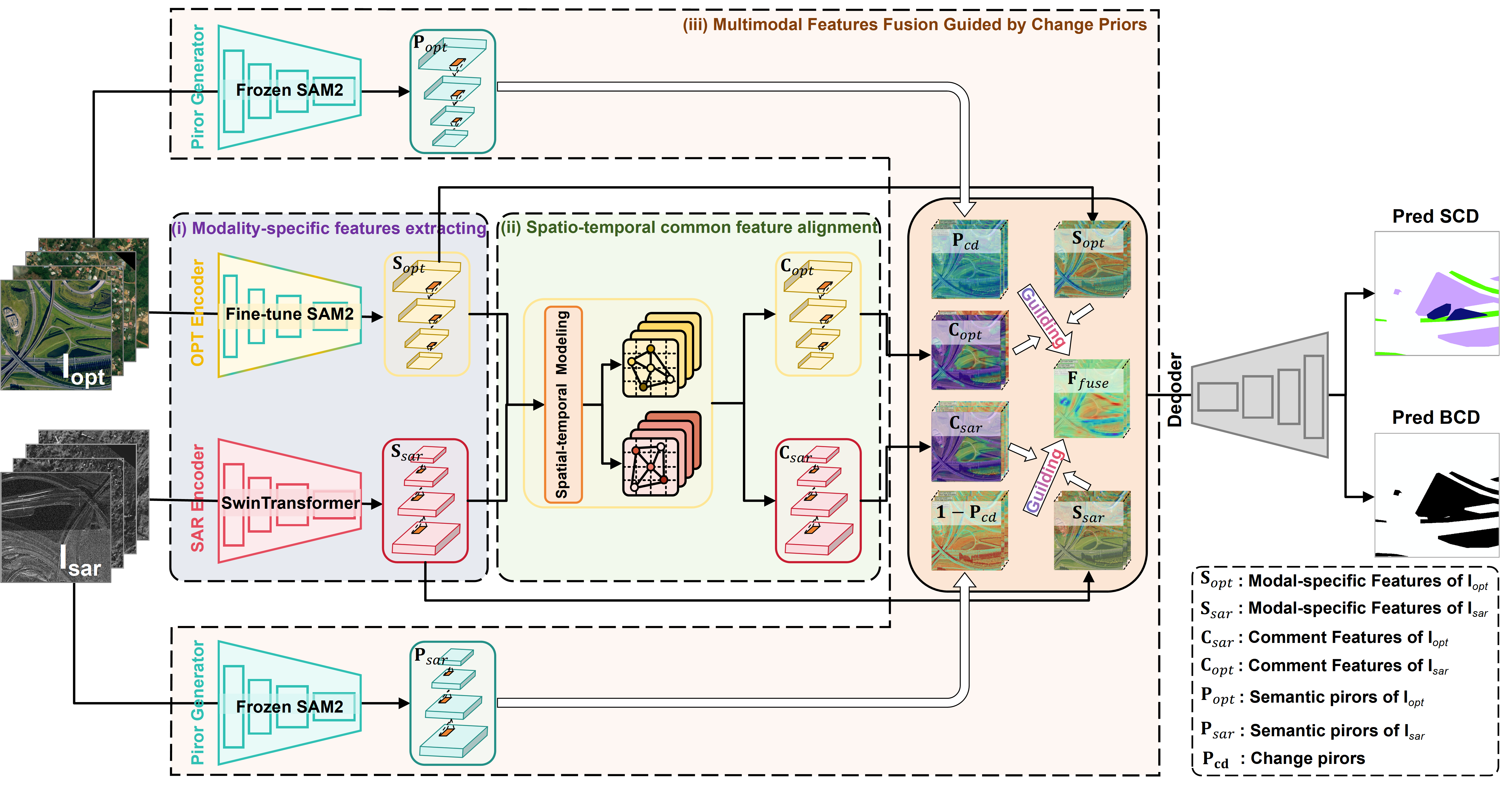}
    \caption{An overview of the proposed OSCD architecture. It contains three core steps: (i) modality-specific feature extraction, (ii) spatio-temporal common feature alignment and (iii) multimodal feature fusion guided by change priors.}
    \label{fig2}
\end{figure*}

We present STSF-Net, a framework designed to synergistically leverage both modality-specific and modality-common features. The overall framework of STSF-Net is shown in Fig. \ref{fig2}. STSF-Net consists of three modules, i.e., Modal-Specific Feature Encoder (MSFE), Spatio-Temporal Common Feature Modeling (STCFM) and Prior-Guided Feature Fusion Module (PGFFM).

Let $\mathbf{I}={\mathbf{I}_{opt},\mathbf{I}_{sar}}$ denote a pair of co-registered optical and SAR images. STSF-Net learns a composite mapping $\phi$ that transforms them into a fused feature space $\mathcal{P}$ suitable for OSCD. The mapping $\phi: \mathcal{I}\to \mathcal{P}$ consists of three consecutive parameterized submodules, i.e., MSFE, STCFM and PGFFM. Specifically, we use MSFE with a pseudo-siamese structure to model multi-scale modality-specific features $\mathbf{S}_m^i$ ($m \in \{\text{opt}, \text{sar}\}$) from $\mathbf{I}_{opt}$ and $\mathbf{I}_{sar}$, where $m$ and $i$ represent the modality type and feature scale, respectively. MSFE preserves the distinctive information inherent in each modality. Subsequently, STCFM is used to learn structure-invariant and temporally consistent representations $\mathbf{C}_m^i$. Next, we employ SAM2 as a frozen semantic prior generator to obtain stable semantic anchors and change priors $\mathbf{P}_{cd}^i$. The multimodal fused feature representation $\mathbf{F}_{fuse}^i$ is generated by adaptively fusing the specific and common features under the guidance of the change priors $\mathbf{P}_{cd}^i$ in the PGFFM. This prior guidance enables the alignment of modality-invariant semantics while leveraging modality-specific characteristics, thereby strengthening discrimination between genuine semantic changes and subtle variations. Finally, $\mathbf{F}_{fuse}^i$ is processed sequentially by a decoder and a classifier to output a dense semantic change probability map.

\subsection{Modality-specific feature extraction}
\begin{figure}[h]
    \centering
    \includegraphics[width=\linewidth]{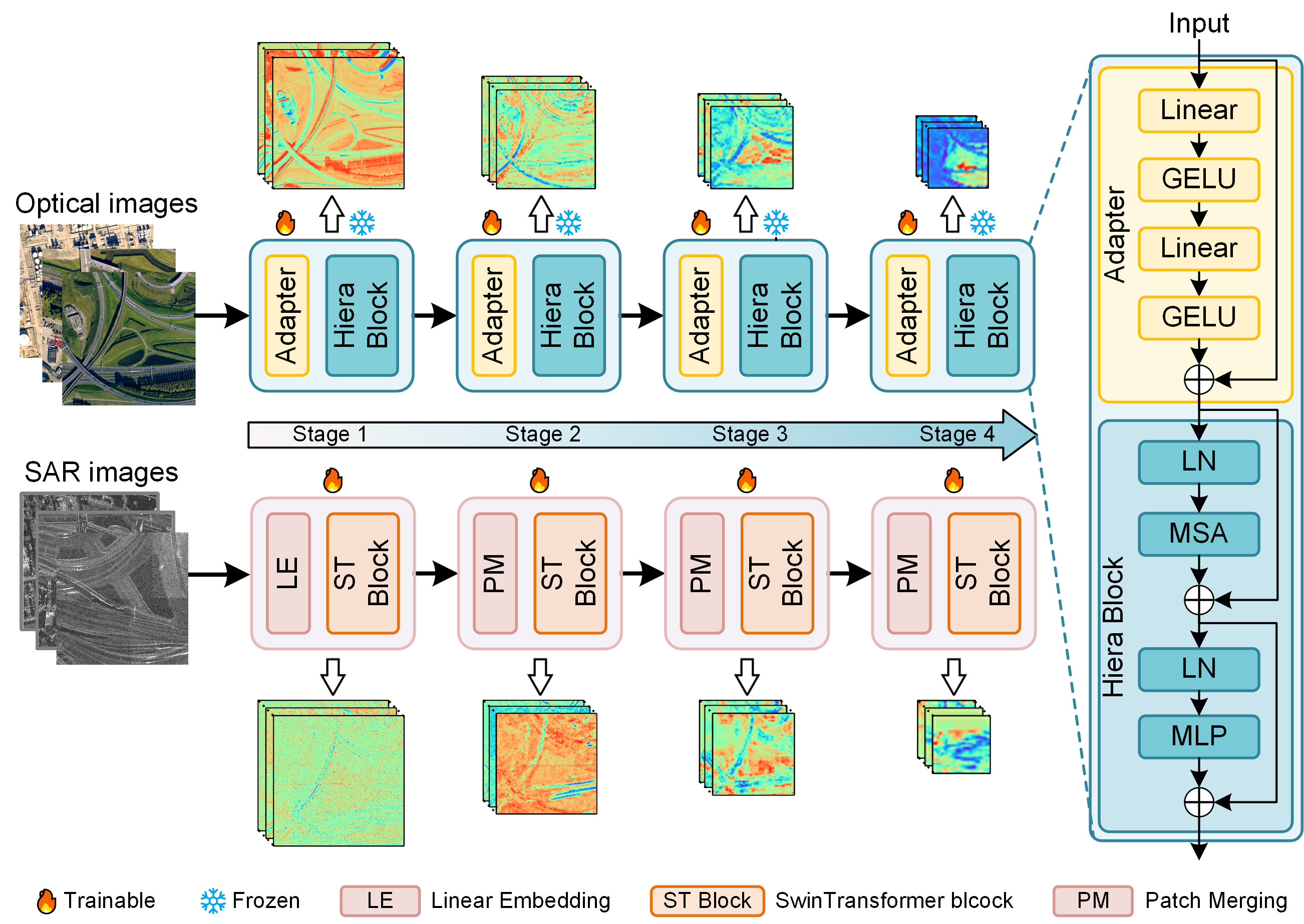}
    \caption{Details of the Modal-specific feature Extractor.}
    \label{fig3}
\end{figure}

To preserve the distinct characteristics of each modality and avoid inter-modal interference, we employ the MSFE with a pseudo-siamese network structure to adaptively extract modality-specific features. As illustrated in Fig. \ref{fig3}, MSFE comprises two distinct branches: an optical encoder and a SAR encoder. Specifically, the optical feature encoder is constructed by fine-tuning SAM2, adapting its general visual priors to the RS context. Meanwhile, for the SAR branch, a small SwinTransformer model is trained from scratch to enable flexible learning of the unique scattering characteristics and structural representations inherent in SAR imagery. 

Each feature encoder contains four encoding stages. In the optical encoder, each stage includes several stacked Adapter and Hiera block units. The Adapter is a lightweight bottleneck module composed of two linear layers with GELU activations. Only the Adapter parameters are updated during the training phase, while the rest of the encoder remains frozen, enabling parameter-efficient fine-tuning. The output of the Adapter is then processed by a Hiera block, a transformer-based module that integrates multi-head self-attention and multi-layer perceptrons. The self-attention mechanism allows interactions between any two positions in the feature map, thus effectively modeling global context. Meanwhile, the SAR feature encoder comprises [2, 2, 6, 2] SwinTransformer blocks across its four stages, enabling the extraction of hierarchical scattering and structural representations. Finally, bi-temporal optical and SAR inputs $\mathbf{I}_{opt}$ and $\mathbf{I}_{sar}$ are processed through Stage 1-4 to generate four scales of bi-temporal semantic features, denoted as $\mathbf{S}_{opt}^{i}$ and $\mathbf{S}_{sar}^{i} \in \mathbb{R}^{C \times H \times W} \quad(i \in {1, 2, 3, 4})$.

\subsection{Spatio-temporal common feature alignment}
\begin{figure*}[!h]
    \centering
    \includegraphics[width=\linewidth]{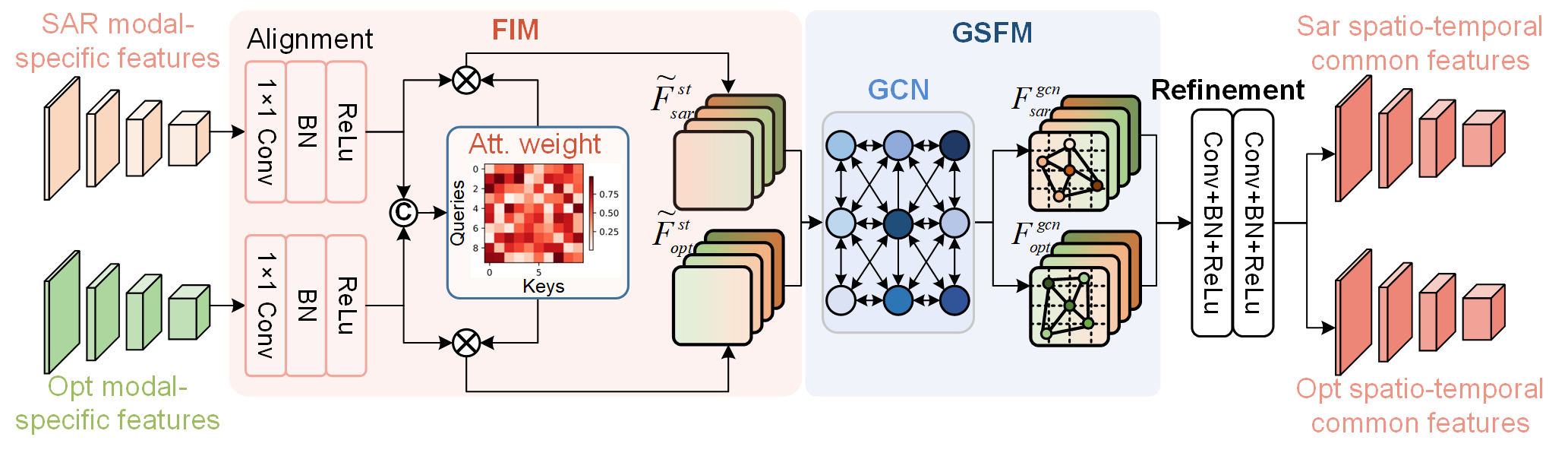}
    \caption{Extraction of cross-modal common features with spatio-temporal correlations.}
    \label{fig4}
\end{figure*}

STCFM is designed to exploit consistent representations across the temporal and spatial dimensions from cross-modal features. While optical and SAR images originate from different imaging mechanisms, the spatial distribution pattern of ground objects exhibits inherent spatio-temporal consistency. By modeling spatio-temporal commonalities, the model is enabled to focus on genuine semantic changes while suppressing pseudo-changes induced by modality-specific noise. Fig. \ref{fig4} shows the details of STCFM. It consists of a cross-modal feature interaction module (FIM) and a Graph Structure Feature Modeling (GSFM) module. The FIM captures temporal dependencies through a cross-attention mechanism, while the GSFM explicitly encodes spatial dependencies through graph convolutions. Together, they embed feature representations that are coherent across both spatial structure and temporal dynamics.

The first stage models the spatio-temporal common features in the FIM, which captures long-range dependencies and temporal correlations between bi-temporal features. Specifically, $\mathbf{S}_{opt}^{i}$ and $\mathbf{S}_{sar}^{i}$ are first projected into a common feature space through a modality-alignment layer for initial feature alignment. The aligned features are concatenated along the channel dimension and passed through a convolutional projector, generating channel attention weights $\mathbf{A}_{m}^{i}$:
\begin{equation}
\mathbf{A}_{m}^{i} = \sigma\left(\text{BN}\left(\text{Conv}_{3\times3}\left(\left[\mathbf{S}_{opt}^{i}, \mathbf{S}_{sar}^{i}\right]\right)\right)\right) \in [0,1]^{H \times W \times 1}.
\end{equation}
Attention weights $\mathbf{A}_{m}^{i}$ are then applied element-wise to recalibrate and weight the original bi-temporal features across the temporal dimension:
\begin{equation}
\mathbf{C}_m^{st,i} = \mathbf{S}_m^i \odot \mathbf{A}_{m}^{i},
\end{equation}
where $\odot$ represents element-wise product.
In the second stage, the GSFM encodes spatial structural dependencies to enhance the representation of spatial distribution patterns. Specifically, $\mathbf{C}_{opt}^{st,i}$ and $\mathbf{C}_{sar}^{st,i}$ are reshaped into a graph sequence $\mathcal{G} = (\mathcal{V}, \mathcal{E})$, where each spatial position is treated as a node $v_i \in \mathrm{V}$ and $\mathcal{E} \subseteq \mathrm{V} \times \mathrm{V}$ denotes the edge set. Two consecutive graph convolution layers with ReLU activation are employed to propagate structural information between nodes. $\mathbf{C}_m^{st,i}$ is first projected into a new feature space via a linear transformation $\mathbf{W}_m^{i,1}$, followed by neighborhood information aggregation using the adjacency matrix $A$:
\begin{equation}
\dot{\mathbf{C}}_m^{gcn,i} = A \cdot \left( \mathbf{C}_m^{st,i} \cdot W_m^{i,1} \right) + b_m^{i,1}.
\end{equation}

To preserve the spatial details while emphasizing structural differences, a residual connection is introduced:
\begin{equation}
\ddot{\mathbf{C}}_m^{gcn,i} = \dot{\mathbf{C}}_m^{gcn,i} - \mathbf{C}_m^{st,i}.
\end{equation}

The residual feature $\ddot{\mathbf{C}}_m^{gcn,i}$ is then passed through a second graph convolutional layer with weights $W_m^{2}$ and a ReLU activation function $\sigma$, producing the refined feature representation $\mathbf{C}_m^{i} \in \mathbb{R}^{N \times C}$:
\begin{equation}
\mathbf{C}_m^{i} = \sigma \left( A \left( \ddot{\mathbf{C}}_m^{gcn,i} \cdot W_m^{i,2} \right) + b_m^{i,2} \right).
\end{equation}

Since $\mathbf{C}_{opt}^{st,i}$ and $\mathbf{C}_{sar}^{st,i}$ already contain bi-temporal spatio-temporal information from the first stage, message propagation between nodes not only aggregates spatial neighborhood features but also implicitly reasons spatial consistency and temporal variations. The resulting graph sequence is reshaped back to the original spatial layout and forwarded to a convolutional feature refinement layer, which further enhances the spatial invariance and temporal coherence.

\subsection{Multimodal features fusion guided by change priors}
\subsubsection{Bi-temporal semantic priors generation}
We propose a SAM2-based multi-scale semantic priors generator (SPG) to introduce stable, high-level semantic guidance for obtaining change features. SPG removes SAM2's components designed for interactive segmentation (e.g., the prompt encoder and mask decoder), and only retains the hierarchical backbone. It performs progressive downsampling to extract multi-level feature representations.

Specifically, SPG takes an input image of arbitrary size and generates feature maps at four distinct scales. Following the practice in SAM2, we integrate (2, 3, 16, 3) Hiera blocks into the four encoding stages, respectively. To ensure the stability and generalizability of the semantic priors, the parameters of SPG are frozen during training. Multi-scale semantic priors $\mathbf{P}_{opt}^{i}$ and $\mathbf{P}_{sar}^{i} \in \mathbb{R}^{C \times H \times W} \quad(i={1, 2, 3, 4})$ are extracted from the optical and SAR images, respectively.

\subsubsection{Prior-guided feature fusion module}
\begin{figure}[!h]
    \centering
    \includegraphics[width=0.8\linewidth]{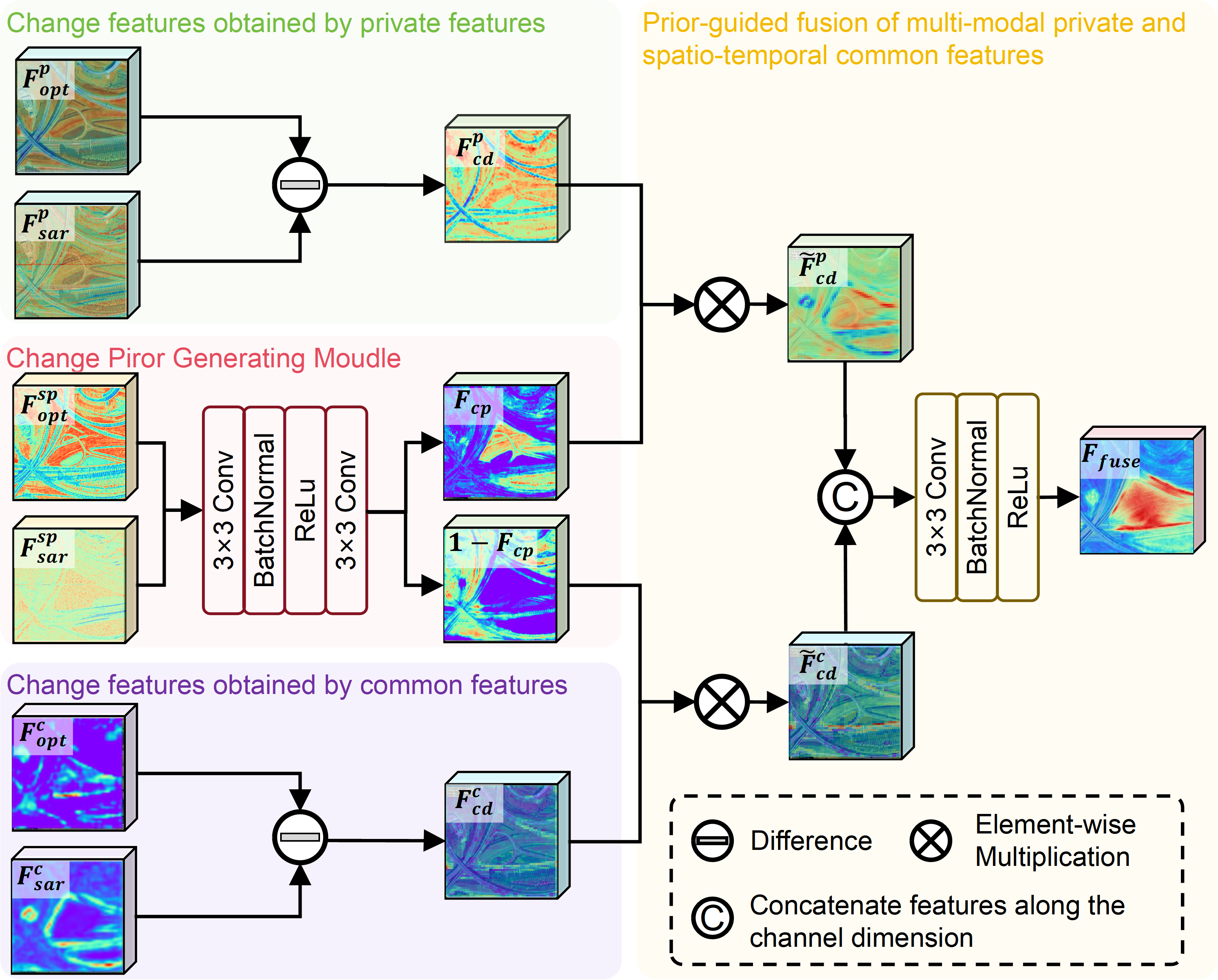}
    \caption{Workflow for fusing modality-specific and common features guided by change priors.}
    \label{fig5}
\end{figure}

We propose a multimodal feature fusion strategy based on change prior guidance, as illustrated in Fig. \ref{fig5}. This strategy utilizes change priors to dynamically adjust the fusion weights between modality-specific change features and common features. In this way, STSF-Net focuses on regions with high change likelihood while preserving structural consistency in unchanged areas.

\textbf{(1) Learning of Change Priors}: We compute multi-scale semantic change priors $\mathbf{P}_{cd}^{i}$ by measuring the per-pixel Euclidean distance between $\mathbf{P}_{opt}^{i}$ and $\mathbf{P}_{sar}^{i}$. The derived feature map highlights candidate change regions. Next, a shallow CNN ($\mathcal{F}$) transforms and enhances the change priors $\mathbf{P}_{cd}^{i}$ at each scale. The enhanced $\mathbf{P}_{cd}^{i}$ is normalized via a Sigmoid activation function to generate a change intensity map $\mathbf{M}_{diff}^{i} \in [0, 1]$, as formulated in Eq.(6):
\begin{equation}
    \mathbf{M}_{diff}^{i} = \sigma(\mathcal{F}(\|\mathbf{P}_{opt}^{i}(i, j, :) - \mathbf{P}_{sar}^{i}(i, j, :)\|_2)). \quad
\end{equation}

$\mathbf{M}_{diff}^{i}$ spatially encodes the prior probability of change at each pixel location, providing a stable guidance signal for the subsequent adaptive fusion of multimodal features. A higher value indicates a greater likelihood that a change has occurred in the region.

\textbf{(2) Prior-Guided Dual-Path Modal Fusion}: We introduce a parallel dual-path fusion strategy to synergistically integrate the strengths of the multi-scale modality-specific and modality-common features. Preliminary change features $\mathbf{F}_{cd}^{si}$ and $\mathbf{F}_{cd}^{ci}$ are first independently extracted from the modality-specific feature space $\mathcal{S}$ and modality-common feature space $\mathcal{C}$ via element-wise subtraction:
\begin{align}
        \mathbf{F}_{cd}^{si} &\in \{\mathbf{S}_{opt}^{i} - \mathbf{S}_{sar}^{i} \mid \mathbf{S}_{opt}^{i}, \mathbf{S}_{sar}^{i} \in \mathcal{S}\}, \nonumber \\
        \mathbf{F}_{cd}^{ci} &\in \{\mathbf{C}_{opt}^{i} - \mathbf{C}_{sar}^{i} \mid \mathbf{C}_{opt}^{i}, \mathbf{C}_{sar}^{i} \in \mathcal{C}\},
\end{align}
where $\mathbf{F}_{cd}^{si}$ captures discrepancies in modality-specific space, which amplifies genuine changes but may be sensitive to modality-dependent noise. In contrast, $\mathbf{F}_{cd}^{ci}$ operates in the aligned common feature space, capturing LCLU semantic changes while retaining structural consistency in unchanged regions, thus being more robust to modal differences. The change probability map $\mathbf{M}_{diff}^{i}$ is then used to weight the fusion of the two paths:
\begin{equation}
\mathbf{F}_{fuse}^{i} = \mathbf{M}_{diff}^{i} \odot \mathbf{F}_{cd}^{si} + (1 - \mathbf{M}_{diff}^{i}) \odot \mathbf{F}_{cd}^{ci},
\end{equation}
where modality-specific features are emphasized in candidate change regions to amplify change saliency. In contrast, common features are weighted by the complementary mask to preserve consistency in probable unchanged areas. The weighted features from both paths are then concatenated and fused through a convolutional module, producing the final fused change representation $\mathbf{F}_{fuse}^{i}$.

This prior-guided fusion mechanism mitigates interference from modal discrepancies, enhances sensitivity to genuine changes, and suppresses false alarms induced by sensor differences.

\section{Datasets}
We conducted experiments on three benchmark OSCD datasets: the Wuhan \citep{ZHANG2022102769}, BRIGHT \citep{essd-BRIGHT}, and the self-constructed Delta-SN6 datasets. 

\begin{figure}[!h]
    \centering
    \includegraphics[width=\linewidth]{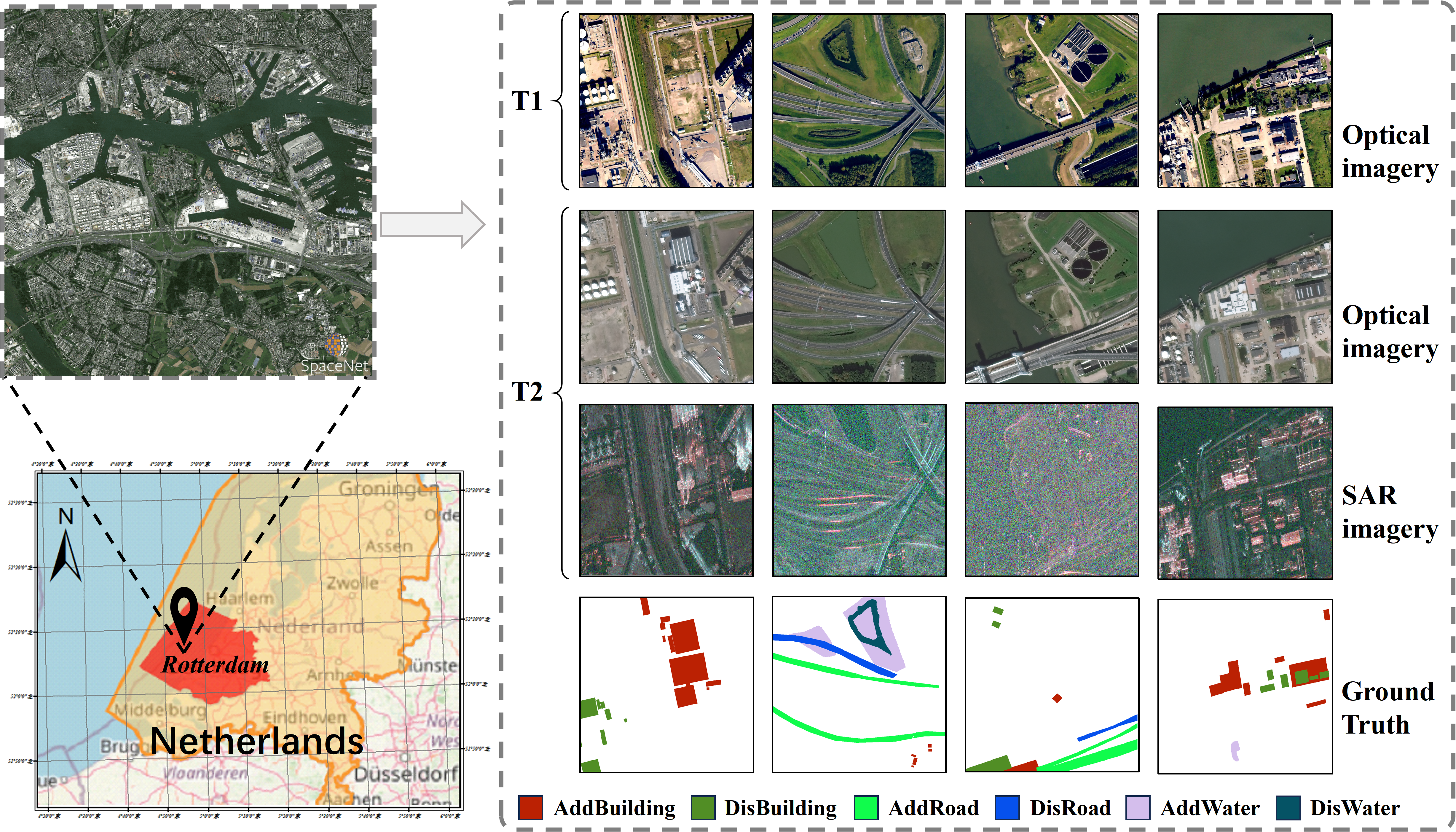}
    \caption{The Delta-SN6 dataset provides multimodal and multi-temporal remote sensing images along with ground truth labels, designed for the OSCD task.}
    \label{fig1}
\end{figure}

Table \ref{tab:dataset_comparison} provides a comparison between the Delta-SN6 and existing OSCD datasets. Delta-SN6 distinctively addresses several key gaps in the field by offering very high-resolution (0.5 m) coregistered optical and fully polarimetric SAR data, along with multiclass, fine-grained change annotations (see Fig. \ref{fig1}). These advantages overcome limitations such as temporal inconsistency, coarse labeling, and limited analytical scope found in prior datasets, thereby providing high-quality, versatile, and task-ready support for advanced OSCD methodologies.

\begin{table}[!ht]
    \centering
    \caption{Comparison of Delta-SN6 with existing OSCD datasets}
    \resizebox{\linewidth}{!}{%
    \begin{tabular}{@{}cccccl@{}}
    \toprule
    \textbf{Dataset} & \textbf{Modality} & \textbf{Resolution} & \textbf{Image Size} & \textbf{Temporal Span} & \textbf{Change Category} \\
    \midrule
    BRIGHT & SAR-optical & 0.3-1 m & 1024×1024 & 2014-2020 & Building \\
    Wuhan & SAR-optical & 2-10 m & 256×256 & 2015-2022 & Building, Road \\
    \rowcolor{lightgray!20}
    Delta-SN6 & \makecell{SAR-optical\\ optical-optical} & 0.5 m & 900×900 & 2007-2019 & Building, Road, Water \\
    \bottomrule
    \end{tabular}}
    \label{tab:dataset_comparison}
\end{table}

Delta-SN6 is built based on the SpaceNet6 dataset \citep{SpaceNet6}. While SpaceNet6 provides high-quality paired SAR and optical images, it lacks the bi-temporal structure essential for CD tasks. To extend its use for OSCD, we have made three key enhancements as follows.

(1) We extended SpaceNet6 to a bi-temporal dataset by adding optical imagery from 2007, extending the observation period to over a decade. This resolves the issue of single-temporal limitation and enables the inclusion of rich change instances.

(2) We provide semantic change labels centered on essential land cover categories, including buildings, roads, and water bodies. Significantly, each change is annotated with its temporal direction, such as road addition or building removal. Unlabeled categories are treated as background during model training and evaluation. In total, the dataset contains 2,818 finely annotated change instances.

(3) Following the SpaceNet6 standard, optical images are segmented according to SAR image strips. Bi-temporal images undergo quality inspection to ensure complete co-registration and adherence to precision requirements for planar errors. Subsequently, the historical images and corresponding annotations are cropped into 900×900-pixel tiles. The dataset is split into 50$\%$ for training, 30$\%$ for testing, and 20$\%$ for validation.

\section{Experiments}
\subsection{Implementation details}
We validate STSF-Net on the three OSCD datasets described in Section IV, and compare it with 13 mainstream CD methods. Comparative methods include CNN-based methods (UNet \citep{UNet}, DeepLabV3+ \citep{Deeplabv3+}, SiamCRNN \citep{SiamCRNN8937755}, SiamAttnUNet \citep{SiamAttnUNet2021132}, DTCDN \citep{LI202114}, M-Unet \citep{9770788}, HGINet \citep{LONG2024318}, HRSICD \citep{WANG2025139}), Transformer-based methods (ChangeOS \citep{ZHENG2021112636}, DamageFormer \citep{Damageformer}, ICIF-Net \citep{ICIF9759285}) and Mamba-based methods (Sigma\citep{wan2025sigma}, GSTM-SCD\citep{LIU202573}, FlowMamba \citep{Guo11299103}). 

Five evaluation metrics are employed to assess the OSCD results, including overall accuracy (OA), Intersection over Union (IoU), Precision, Recall and F1 scores. The $F_1^{bcd}$ and $F_1^{clf}$ focus on the accuracy of change areas and semantic segments, respectively.

All experiments were conducted on a hardware platform equipped with an NVIDIA L20-48GB GPU. To ensure fairness in the experimental comparisons, we used the same datasets, code execution environment and hyperparameter configurations to train all models. During the training of the proposed method, the Adam optimizer was employed, with the number of iterations set to $6\times10^{4}$, a batch size of 8, and an initial learning rate of $5\times10^{-4}$.

\subsection{Comparative experimental results and analysis}
\subsubsection{Results on the Wuhan dataset}
Table \ref{tab:heterogeneous_change_evaluation} reports results on the Wuhan dataset. STSF-Net achieves the highest $F_1^{bcd}$ (57.79$\%$), mIoU (64.25$\%$), OA (88.71$\%$), and Precision (60.57$\%$), improving $F_1^{bcd}$ and mIoU by 1.09$\%$ and 1.30$\%$ over the second-best GSTM-SCD. Although GSTM-SCD attains the highest Recall (58.73$\%$), its lower Precision (54.80$\%$) indicates more false positives.

\begin{figure*}[!h]
    \centering
    \includegraphics[width=\linewidth]{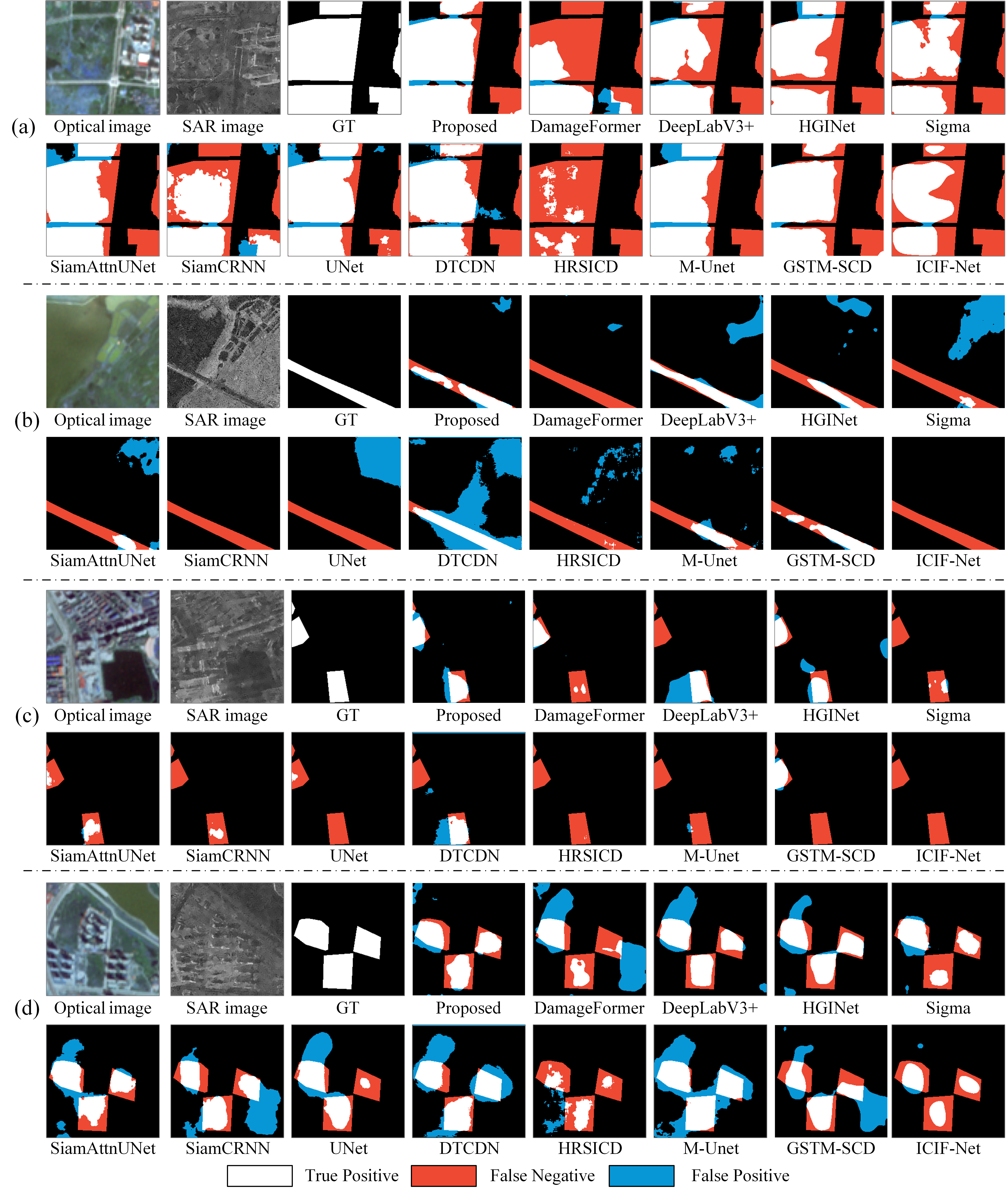}
    \caption{Qualitative results provided by different methods in the testing areas (a)-(d) on the Wuhan dataset.}
    \label{fig6}
\end{figure*}

Fig. \ref{fig6} shows that HRSICD, SiamAttnUNet, and DamageFormer produce fragmented results with substantial false alarms. DeepLabV3+, HGINet, and DTCDN exhibit improved recall but incompletely capture changed regions (Fig. \ref{fig6}(a)-(c)). GSTM-SCD and HGINet struggle with detail preservation and false alarms (Fig. \ref{fig6}(d)). STSF-Net mitigates these issues through heterogeneous feature fusion and spatial context modeling, achieving superior spatial continuity and boundary clarity. Its results align closely with GT, particularly in dense building complexes.

\begin{table}[htbp]
\centering
\caption{Evaluation of OSCD accuracy results ($\%$) on the Wuhan dataset. The best and second-best results are in bold and underlined, respectively.}
\label{tab:heterogeneous_change_evaluation}
\resizebox{0.9\linewidth}{!}{%
\begin{tabular}{lccccccc}
\toprule
Method & Reference & Recall & Precision & OA & $F_1^{bcd}$ & mIoU \\
\midrule
DeepLabV3+ & ECCV'18 & 52.80 & 54.54 & 87.25 & 53.66 & 61.45 \\
SiamCRNN & TGRS'20 & 48.65 & 33.49 & 79.30 & 39.67 & 51.27 \\
SiamAttnUNet & ISPRS'21 & 53.53 & \underline{58.18} & 88.12 & 55.76 & 62.91 \\
DTCDN & ISPRS'21 & \underline{58.03} & 42.90 & 83.32 & 49.33 & 57.30 \\
ChangeOS & RSE'21 & 42.80 & 57.42 & \underline{87.56} & 49.04 & 59.63 \\
ICIF-Net & TGRS'22 & 44.41 & 58.13 & 86.08 & 50.35 & 57.84 \\
M-Unet & GRSL'22 & 52.48 & 52.25 & 87.21 & 52.36 & 37.33 \\
DamageFormer & IGARSS'22 & 48.36 & 39.08 & 82.23 & 43.23 & 54.26 \\
HGINet & ISPRS'24 & 48.80 & 47.04 & 85.16 & 47.91 & 57.78 \\
HRSICD & ISPRS'25 & 18.19 & 51.66 & 86.18 & 26.91 & 15.55 \\
Sigma & WACV'25 & 37.43 & 40.15 & 83.45 & 38.74 & 53.28 \\
GSTM-SCD & ISPRS'25 & \textbf{58.73} &54.80  & 86.58 & \underline{56.70} & \underline{62.95} \\
FlowMamba & TSCVT'25 & 57.18 & 53.11 & 84.89 & 55.07 & 61.57 \\
\rowcolor{lightgray!20}
STSF-Net & \textit{Proposed} & 55.25 & \textbf{60.57} & \textbf{88.71} & \textbf{57.79} & \textbf{64.25} \\
\bottomrule
\end{tabular}
}
\end{table}

\subsubsection{Results on the BRIGHT dataset}
Table \ref{tab:building-damage-evaluation} demonstrates that STSF-Net also obtains the highest metrics with $F_1^{clf}$ of 75.83$\%$ and mIoU of 67.91$\%$. Compared to GSTM-SCD, IoU for damaged and destroyed buildings increases by 1.97$\%$ and 0.94$\%$, respectively. The IoU for destroyed buildings reaches 56.80$\%$, which is 3.82$\%$ higher than that of DamageFormer, highlighting the advantage of STSF-Net in detecting severely damaged structures. Furthermore, the IoU of intact buildings is as high as 75.01$\%$, which is crucial for reducing false alarms. 

\begin{figure*}[!h]
    \centering
    \includegraphics[width=\linewidth]{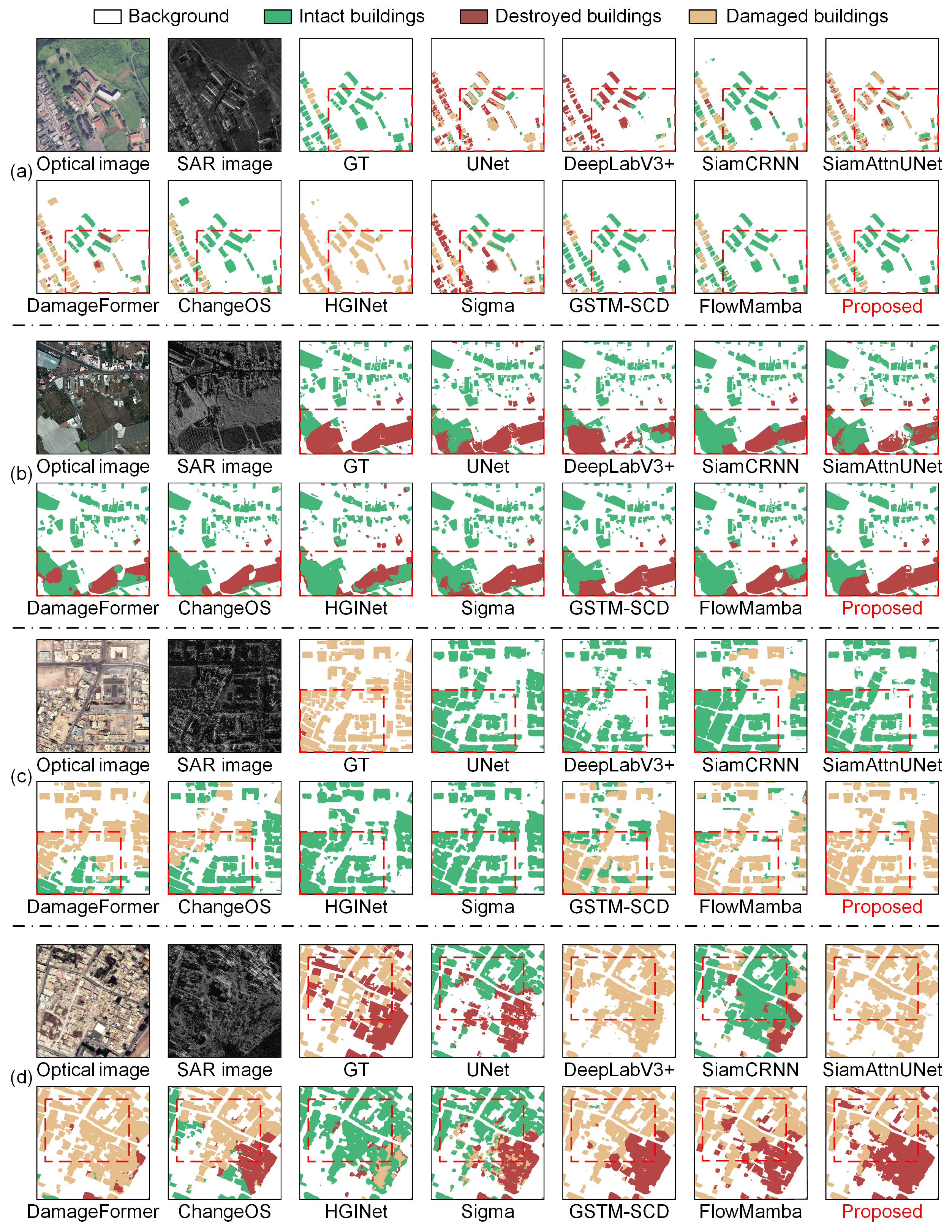}
    \caption{Qualitative results provided by different methods in the testing areas (a)-(d) on the BRIGHT dataset.}
    \label{fig7}
\end{figure*}

As displayed in Fig. \ref{fig7}, STSF-Net effectively decreases building omissions and demonstrates the highest accuracy in recognizing damaged and destroyed buildings. In complex scenes, particularly in areas where buildings exhibit varying degrees of damage and are sparsely distributed (Fig. \ref{fig7}(d)), STSF-Net exhibits distinct advantages in terms of completeness of change regions and semantic segment accuracy. HGINet and DeepLabV3+ display significant omission and misclassification errors on damaged buildings (Fig. \ref{fig7}(c)). In contrast, the results obtained by STSF-Net exhibit a higher level of agreement with the GT, particularly for damaged and destroyed buildings (Fig. \ref{fig7}(b)). Results of UNet and Sigma produce blurred boundaries, whereas our method maintains better structural integrity (Fig. \ref{fig7}(a)). Although ChangeOS and SiamCRNN exhibit improved shape consistency with GT, they still suffer from a serious problem of misclassifying damaged buildings as intact ones. GSTM-SCD and DamageFormer also encounter similar issues.

\begin{table}[htbp]
\centering
\caption{Evaluation of building damage accuracy results ($\%$) of different methods on the BRIGHT dataset.}
\label{tab:building-damage-evaluation}
\resizebox{\linewidth}{!}{%
\begin{tabular}{lcccccccc}
\toprule
\multirow{2}{*}{Methods} & \multirow{2}{*}{$F_1^{bcd}$} & \multirow{2}{*}{$F_1^{clf}$} & \multirow{2}{*}{OA} & \multirow{2}{*}{mIoU} & \multicolumn{4}{c}{IoU per class} \\
\cmidrule(lr){6-9}
& & & & & {Background} & {Intact} & {Damaged} & {Destroyed} \\
\midrule
UNet & 86.01 & 62.94 & 94.60 & 59.22 & 95.46 & 67.35 & 28.44 & 45.63 \\
DeepLabV3+ & 82.61 & 70.33 & 93.60 & 59.37 & 94.82 & 63.96 & 32.19 & 46.53 \\
SiamCRNN & 88.77 & 68.71 & 95.42 & 63.37 & 96.32 & 71.45 & 35.06 & 50.67 \\
SiamAttnUNet & 87.21 & 70.22 & 94.99 & 62.82 & 95.83 & 69.45 & 36.03 & 49.96 \\
ChangeOS & 89.60 & 71.88 & 95.84 & 65.98 & 96.54 & 73.85 & 38.99 & 54.53 \\
DamageFormer & 89.92 & 72.22 & 95.95 & 66.09 & \underline{96.75} & 74.30 & 40.35 & 52.98 \\
HGINet & 83.89 & 30.58 & 94.17 & 50.70 & 95.18 & 65.76 & 7.76 & 34.08 \\
Sigma & 86.82 & 64.20 & 95.16 & 62.31 & 95.86 & 70.15 & 35.27 & 47.95 \\
GSTM-SCD & \underline{90.51} & 73.96 & 95.88 & 66.83 & 96.63 & 73.75 & 41.09 & 55.86 \\
FlowMamba &89.45  &\underline{74.18}  &\underline{96.15}  &\underline{67.04}  &96.20  & \underline{74.88} & \underline{42.33} & \underline{56.41} \\
\rowcolor{lightgray!20}
STSF-Net & \textbf{91.71} & \textbf{75.83} & \textbf{96.10} & \textbf{67.91} & \textbf{96.78} & \textbf{75.01} & \textbf{43.06} & \textbf{56.80} \\
\bottomrule
\end{tabular}%
}
\end{table}

\subsubsection{Results on the Delta-SN6 dataset}
Table \ref{tab:sn6-hscd-evaluation} demonstrates that the proposed method achieves the best performance across all evaluation metrics, with an $F_1^{bcd}$ of 94.60\%, $F_1^{clf}$ of 95.27\%, OA of 99.54\%, and mIoU of 91.33\%. It represents a significant improvement over existing methods, with gains of 3.18\% in $F_1^{bcd}$ over GSTM-SCD and 3.21\% in mIoU over DamageFormer. Notably, STSF-Net achieves the highest accuracy in per-class IoU across all change categories, further confirming STSF-Net's superior discriminative capability among different change classes.

\begin{figure*}[!h]
    \centering
    \includegraphics[width=\linewidth]{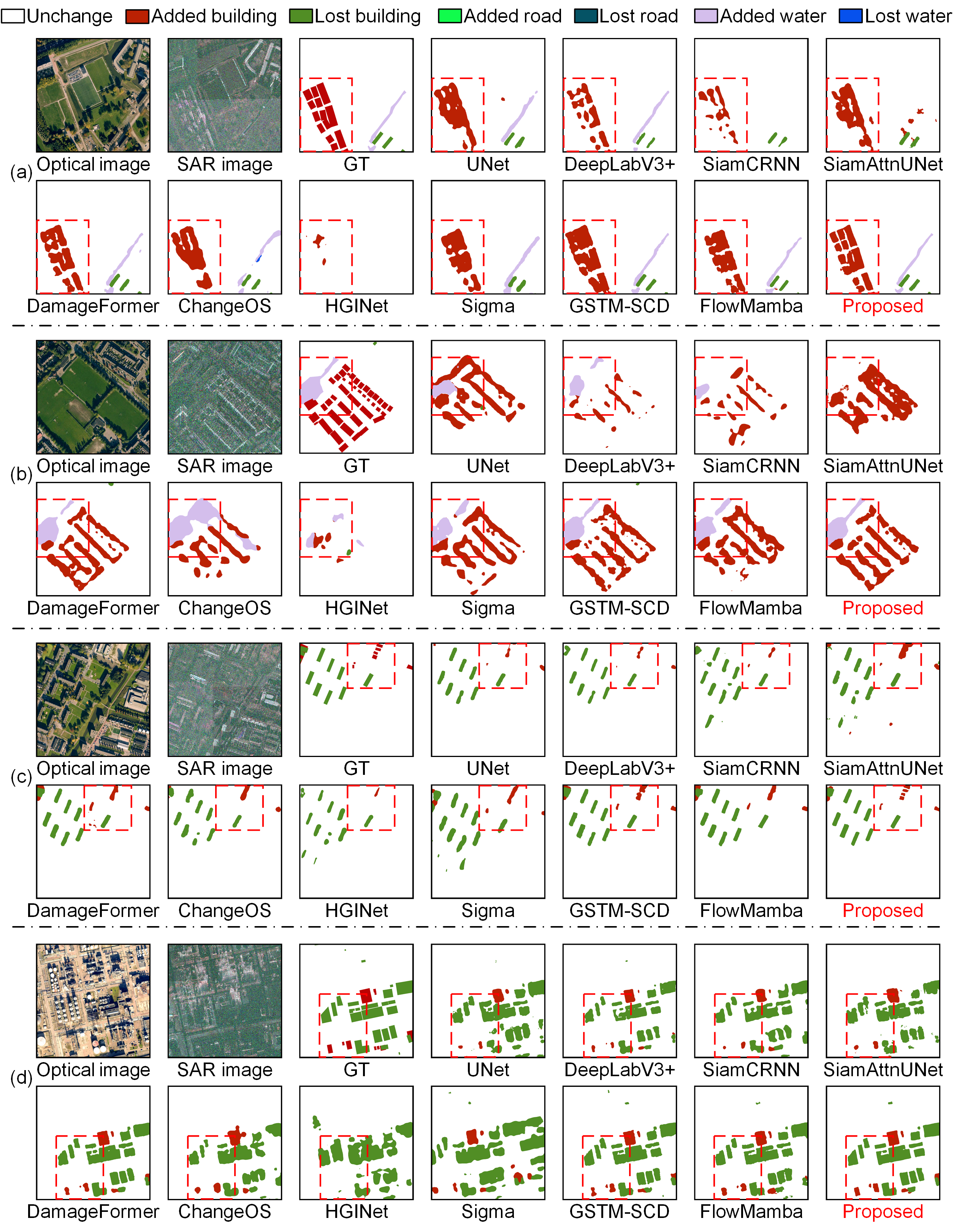}
    \caption{Qualitative results provided by different methods in the testing areas (a)-(d) on the Delta-SN6 dataset.}
    \label{fig8}
\end{figure*}

Visualization results (Fig. \ref{fig8}) further elucidate performance disparities. For new construction detection, STSF-Net exhibits high consistency with GT, while DamageFormer and GSTM-SCD show edge blurring artifacts. DeepLabV3+ and HGINet demonstrate significant omission errors, as evidenced by undetected new buildings in the testing areas (a) and (b). In changed road detection, STSF-Net maintains sharp boundary delineation, whereas SiamCRNN and ChangeOS exhibit false positive detections. For mixed-change scenes (e.g., the testing area (b) with added water and disappeared buildings), STSF-Net effectively distinguishes these classes, avoiding the confusion observed in SiamAttnUNet and UNet (see testing areas (c) and (d)).

\begin{table}[htbp]
\centering
\caption{Evaluation of OSCD accuracy results ($\%$) on the Delta-SN6 dataset. BG, AB, AR, AW, DB, DR and DW refer to background, added building, added road, added water, disappeared building, disappeared road and disappeared water, respectively.}
\label{tab:sn6-hscd-evaluation}
\resizebox{\linewidth}{!}{%
\begin{tabular}{@{}lcccccccccccc@{}}
\toprule
\multirow{2}{*}{Methods} & \multirow{2}{*}{$F_1^{bcd}$} & \multirow{2}{*}{$F_1^{clf}$} & \multirow{2}{*}{OA} & \multirow{2}{*}{mIoU} & \multicolumn{7}{c}{IoU per class} \\
\cmidrule(lr){6-12}
& & & & & BG & AB & AR & AW & DB & DR & DW \\
\midrule
UNet & 83.28 & 26.37 & 98.53 & 53.44 & 98.54 & 62.10 & 57.89 & 77.47 & 78.09 & 0.00 & 0.00 \\
DeepLabV3+ & 90.83 & 22.32 & 98.58 & 51.88 & 98.58 & 64.21 & 29.04 & 73.57 & 78.87 & 2.95 & 15.95 \\
SiamCRNN & 82.17 & 54.20 & 98.36 & 57.04 & 98.38 & 52.83 & 28.25 & 70.20 & 78.83 & 15.09 & 55.73 \\
SiamAttnUNet & 82.80 & 88.15 & 98.54 & 74.19 & 98.53 & 61.81 & 54.10 & 74.34 & 79.37 & 74.95 & 76.20 \\
ChangeOS & 78.57 & 86.53 & 98.10 & 66.92 & 98.13 & 55.83 & 43.02 & 67.59 & 71.69 & 61.16 & 71.01 \\
DamageFormer & 91.38 & \underline{94.81} & \underline{99.37} & \underline{88.12} & 99.04 & 80.44 & \underline{75.96} & \underline{91.42} & \underline{90.98} &88.75 & \underline{90.26} \\
HGINet & 61.09 & 19.78 & 96.81 & 26.21 & 96.88 & 23.87 & 0.73 & 13.79 & 48.22 & 0.00 & 0.00 \\
Sigma & 78.23 & 91.39 & 98.20 & 70.94 & 98.21 & 57.16 & 50.93 & 75.28 & 71.03 & 67.93 & 76.03 \\
GSTM-SCD & \underline{91.42} & 94.61 & 99.32 & 87.47 & 99.32 & \underline{80.88} & 73.95 & 89.52 & 90.60 & 88.36 & 89.67 \\
FlowMamba & 91.32 & 94.56 & 99.30 & 87.23 & 99.30 & 79.60 & 72.58 & 90.00 & 89.90 & \underline{89.38} & 89.37 \\
\rowcolor{lightgray!20}
STSF-Net & \textbf{94.60} & \textbf{95.27} & \textbf{99.54} & \textbf{91.33} & \textbf{99.55} & \textbf{85.68} & \textbf{81.42} & \textbf{92.95} & \textbf{92.77} & \textbf{94.62} & \textbf{92.33} \\
\bottomrule
\end{tabular}}
\end{table}

\subsection{Ablation study}
\subsubsection{Effectiveness of the proposed modules}
We conducted ablation studies on BRIGHT and Delta-SN6 to validate each proposed module. An asymmetric Siamese network (SAM2 and Swin Transformer encoders) serves as the baseline; all eight configurations of FIM, GSFM, and PGFFM are evaluated in Table \ref{tab:ablation-study}.

\begin{figure}[!h]
    \centering
    \includegraphics[width=\linewidth]{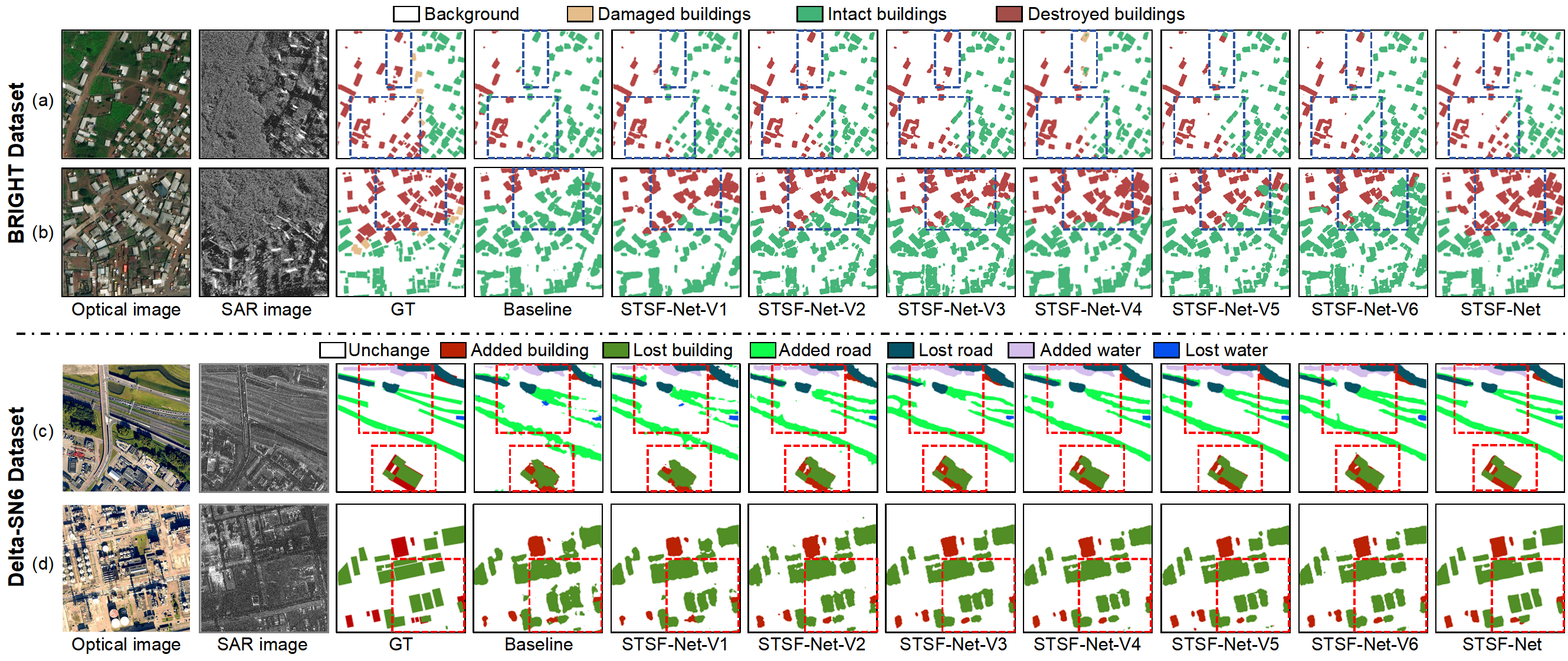}
    \caption{Qualitative comparison of the OSCD results obtained using different network modules on the BRIGHT and Delta-SN6 datasets. STSF-Net-V1: baseline+FIM, STSF-Net-V2: baseline+GSFM, STSF-Net-V3: baseline+PGFFM, STSF-Net-V4: baseline+FIM+GSFM, STSF-Net-V5: baseline+FIM+PGFFM, STSF-Net-V6: baseline+GSFM+PGFFM.}
    \label{fig9}
\end{figure}

\textbf{(1) Progressive Integration.} Performance steadily improves with progressive module incorporation (Table \ref{tab:ablation-study} and Fig.~\ref{fig9}). STSF-Net-V1 (FIM only) improves mIoU by 0.77$\%$ and 3.03$\%$ on BRIGHT and Delta-SN6, though damaged building classification remains biased (Fig.~\ref{fig9}(a)). Adding GSFM (STSF-Net-V4) improves discrimination of damaged versus intact buildings (Figs.~\ref{fig9}(b)-(d)), reaching mIoU of 65.64\% and 88.21\%. The full STSF-Net achieves optimal mIoU of 67.91\% and 91.33\%, confirming module complementarity.

\textbf{(2) Single-module Analysis.} Single-module variants consistently rank V3 $>$ V2 $>$ V1 by mIoU on both datasets. PGFFM contributes most strongly via VFM semantic priors, GSFM provides the second-strongest contribution through spatial structure modeling, and FIM is the most modest. All outperform the baseline. Fig.~\ref{fig9} reveals that V1 misclassifies damaged buildings and exhibits omissions, indicating that feature alignment alone is insufficient without spatial reasoning and prior guidance.

\textbf{(3) Module Combinations.} Two-module combinations surpass single-module counterparts. The full model exceeds the best two-module variant by 0.76\% (BRIGHT) and 2.51\% (Delta-SN6) in mIoU, confirming that FIM, GSFM, and PGFFM each provide indispensable representations whose synergy establishes a robust OSCD framework.

\begin{table}[htbp]
\centering
\caption{Ablation study results ($\%$) of the proposed modules on BRIGHT and Delta-SN6 datasets.}
\label{tab:ablation-study}
\resizebox{\linewidth}{!}{%
\begin{tabular}{@{}lcccccccccccc@{}}
\toprule
\multirow{2}{*}{Methods} & \multicolumn{3}{c}{Proposed Module} & \multicolumn{4}{c}{BRIGHT Dataset (\%)} & \multicolumn{4}{c}{Delta-SN6 Dataset (\%)} \\
\cmidrule(lr){2-4} \cmidrule(lr){5-8} \cmidrule(lr){9-12}
& FIM & GSFM & PGFFM & F1$_{bcd}$ & F1$_{clf}$ & OA & mIoU & F1$_{bcd}$ & F1$_{clf}$ & OA & mIoU \\
\midrule
Baseline & $\times$ & $\times$ & $\times$ & 88.14 & 70.87 & 95.39 & 63.99 & 79.83 & 87.46 & 98.34 & 71.42 \\
STSF-Net-V1 & $\checkmark$ & $\times$ & $\times$ & 88.51 & 71.81 & 95.44 & 64.76 & 81.63 & 87.38 & 98.49 & 74.45 \\
STSF-Net-V2 & $\times$ & $\checkmark$ & $\times$ & 88.85 & 72.21 & 95.73 & 65.52 & 89.75 & 90.88 & 99.17 & 85.87 \\
STSF-Net-V3 & $\times$ & $\times$ & $\checkmark$ & 88.21 & 74.54 & 95.43 & 66.28 & 91.82 & 91.91 & 99.32 & 87.45 \\
STSF-Net-V4 & $\checkmark$ & $\checkmark$ & $\times$ & 88.63 & 73.32 & 95.69 & 65.64 & 92.40 & 92.42 & 99.35 & 88.21 \\
STSF-Net-V5 & $\checkmark$ & $\times$ & $\checkmark$ & 88.95 & 74.92 & 95.56 & 67.15 & 92.39 & 92.72 & 99.32 & 88.44 \\
STSF-Net-V6 & $\times$ & $\checkmark$ & $\checkmark$ & 89.51 & 74.54 & 95.66 & 66.96 & 93.11 & 93.16 &99.41 & 88.82 \\
\rowcolor{lightgray!20}
STSF-Net & $\checkmark$ & $\checkmark$ & $\checkmark$ & \textbf{91.71} & \textbf{75.83} & \textbf{96.10} & \textbf{67.91} & \textbf{94.60} & \textbf{95.27} & \textbf{99.54} & \textbf{91.33} \\
\bottomrule
\end{tabular}%
}
\end{table}

\subsubsection{Effectiveness of the MSFE}
We validate the effectiveness of the proposed MSFE based on SAM2 on the Delta-SN6 dataset. Specifically, we replaced the proposed encoder with other architectures, including the Siamese SwinTransformer encoder and the Siamese SAM2 encoder, within the STSF-Net framework. As shown in Table \ref{tab:model_comparison}, the proposed MSFE outperforms the comparative encoders across all evaluation metrics, achieving an $F_1^{bcd}$ of 94.60$\%$, an mIoU of 91.33$\%$, and an $F_1^{clf}$ of 95.27$\%$. This advantage was particularly evident in per-class IoU metrics, especially for added buildings (85.68$\%$), added roads (81.42$\%$), and disappeared roads (94.62$\%$). It demonstrates that the MSFE effectively leverages SAM2's general visual priors to adapt to optical imagery, while its SAR feature extractor captures scattering characteristics and structural patterns.

\begin{table*}[htbp]
\centering
\caption{Performance comparison of different encoder architectures on the Delta-SN6 dataset.}
\label{tab:model_comparison}
\resizebox{\linewidth}{!}{%
\begin{tabular}{@{}lccccccccccccr@{}}
\toprule
\multirow{2}{*}{Methods} & 
\multirow{2}{*}{\makecell{Trainable\\Params (M)}} &
\multirow{2}{*}{$F_1^{bcd}$} & 
\multirow{2}{*}{$F_1^{clf}$} & 
\multirow{2}{*}{OA} & 
\multirow{2}{*}{mIoU} & 
\multicolumn{7}{c}{IoU per class} &  \\
\cmidrule(lr){7-13}
& & & & & & BG & AB & AR & AW & DB & DR & DW  \\
\midrule
Siamese SwinTransformer & 56.58 & \underline{91.76} & \underline{95.05} & \underline{99.31} & \underline{87.82} & \underline{99.31} & \underline{78.37} & \underline{76.14} & \underline{91.26} & \underline{89.28} & \underline{90.15} & \underline{90.24}  \\
Siamese SAM2 & \textbf{0.72} & 90.87 & 94.02 & 99.23 & 85.84 & 99.24 & 77.41 & 72.53 & 89.17 & 87.62 & 86.96 & 87.95  \\
\rowcolor{lightgray!20}
SAM2 + SwinTransformer & \underline{29.01} & \textbf{94.60} & \textbf{95.27} & \textbf{99.54} & \textbf{91.33} & \textbf{99.55} & \textbf{85.68} & \textbf{81.42} & \textbf{92.95} & \textbf{92.77} & \textbf{94.62} & \textbf{92.33}  \\
\bottomrule
\end{tabular}}
\end{table*}

\subsubsection{Effectiveness of the STCFM}
\begin{figure}[!h]
    \centering
    \includegraphics[width=\linewidth]{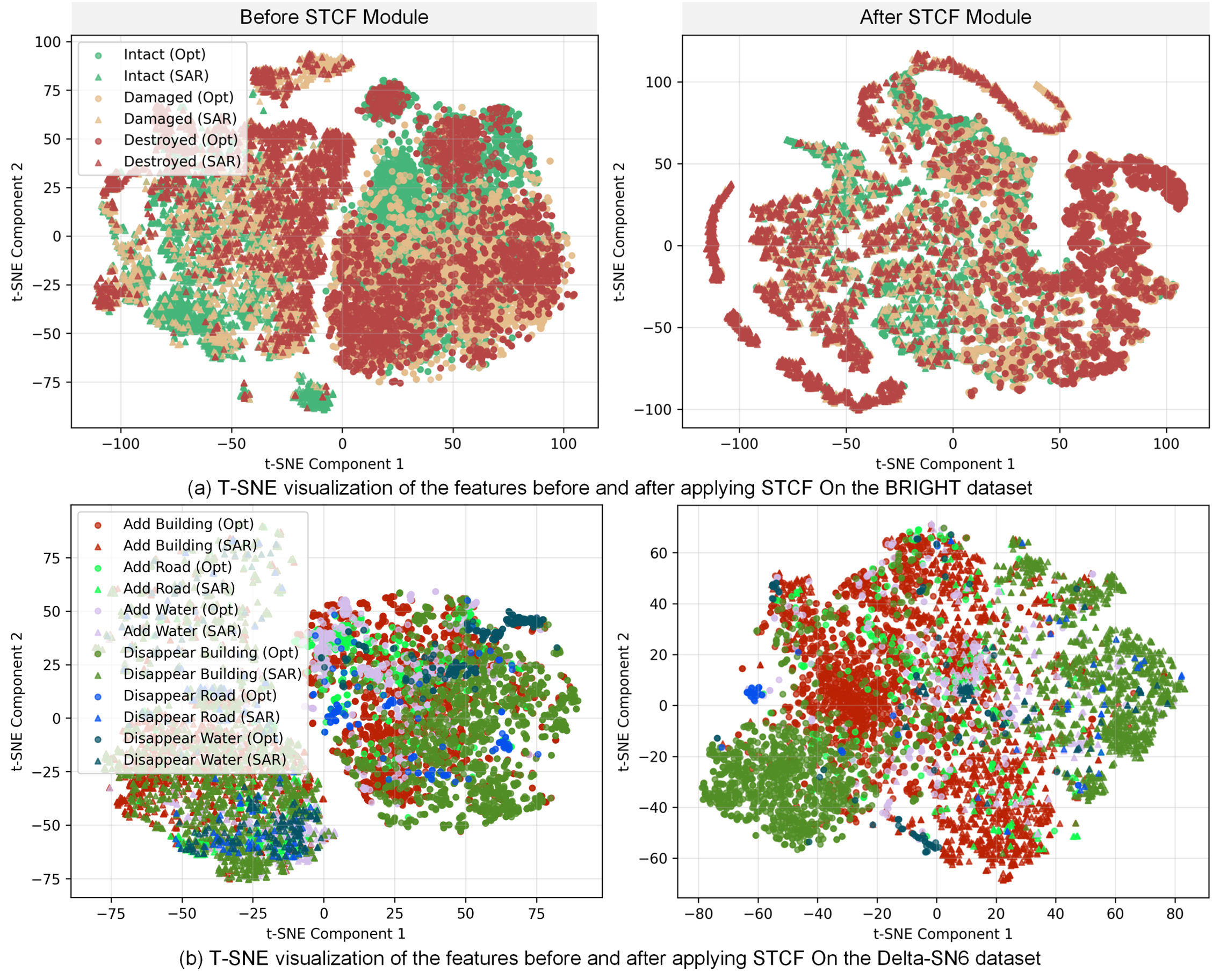}
    \caption{Comparison of feature distributions of optical and SAR before and after STCFM processing.}
    \label{fig10}
\end{figure}

We leveraged t-distributed stochastic neighbor embedding (t-SNE) to visualize the optical and SAR feature distributions, both before and after the integration of the STCFM. As shown in Fig.~\ref{fig10}, before integrating the STCFM module, the optical and SAR features in the feature space exhibit distinct separation, forming well-defined independent clusters. This distribution indicates a significant modality gap, which prevents the model from learning modality-common features and effectively fusing complementary information. After applying the STCFM module, multimodal features are effectively aligned. Specifically, the interweaving of optical and SAR feature points within the same semantic category results in more tightly integrated mixed clusters. This demonstrates that STCFM successfully facilitates cross-modal interaction and learns more consistent and modality-agnostic feature representations. In addition, STCFM also enhances intra-class compactness. Feature points belonging to the same change type, such as buildings and roads, exhibit tighter clustering after fusion, which suggests that STCFM learns more robust and generalizable change semantic representations.

\subsubsection{Effectiveness of the PGFFM}
To validate the PGFFM, we compared different VFMs used as change-prior generators. Specifically, we evaluated STSF-Net performance with and without prior guidance, and with SAM, SAM2, and DINOv3 as change-prior generators, respectively. As shown in Table \ref{tab:performance_comparison}, STSF-Net without PGFFM achieves an mIoU of 65.64$\%$ and 84.93$\%$ on the BRIGHT and Delta-SN6 datasets, respectively. When integrating PGFFM with different prior generators, i.e., SAM, SAM2, and DINOv3, the model consistently demonstrates improved performance, with mIoU reaching 88.16$\%$, 91.33$\%$, and 88.51$\%$ on Delta-SN6, respectively. Among them, DINOv3 slightly outperforms SAM (88.51$\%$ vs.\ 88.16$\%$), though both trail behind SAM2. This trend holds on the BRIGHT dataset as well, where the SAM2-based configuration achieves the highest mIoU of 67.91$\%$. The widespread performance improvements validate the effectiveness of the prior-guided fusion mechanism itself, rather than fortuitous advantages from specific visual models. Notably, the gains vary across different prior generators, with SAM2-based PGFFM yielding the best results due to its strong general segmentation capabilities and precise modeling of object boundaries, highlighting that prior quality is a critical factor influencing fusion performance.

\begin{figure}[!h]
    \centering
    \includegraphics[width=\linewidth]{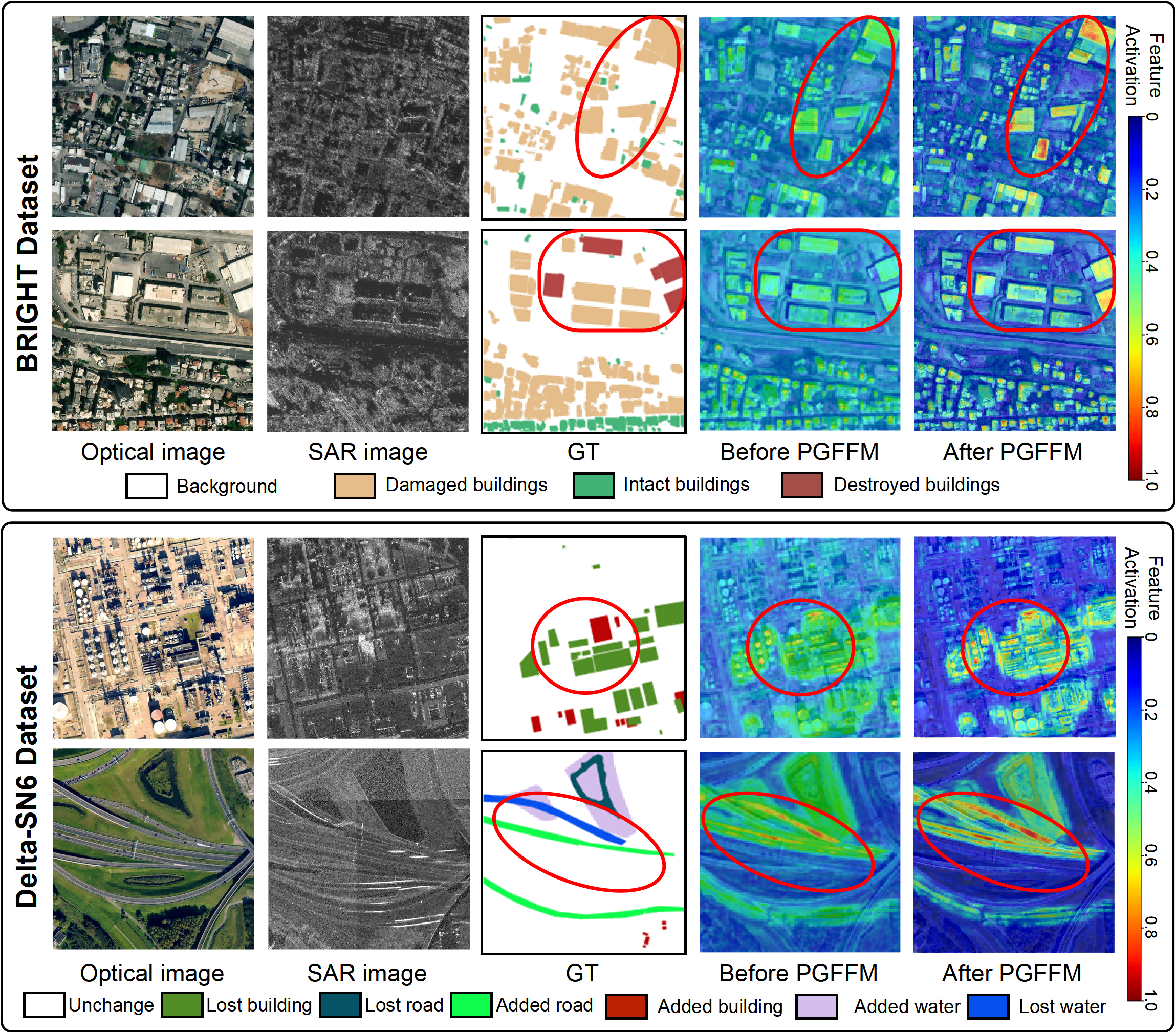}
    \caption{Visualized change in feature response intensity before and after PGFFM addition on BRIGHT and Delta-SN6 datasets.}
    \label{fig11}
\end{figure}

Qualitative visualizations (see Fig. \ref{fig11}) provide intuitive attribution for these improvements. Without prior guidance, feature response maps exhibit weak contrast between changed and unchanged regions, with weak activation in changed areas, making it difficult for the model to accurately localize change boundaries. After integrating PGFFM (particularly with SAM2 as the prior), activation intensity in true change regions is significantly enhanced, while activation in unchanged background regions is effectively suppressed, improving inter-class separability. The visual evidence demonstrates that PGFFM does not simply amplify all features but adaptively enhances discriminative features related to change semantics based on the prior, guiding the model to focus on essential differences.

\begin{table}[htbp]
\centering
\caption{Performance comparison of different methods on BRIGHT and Delta-SN6 datasets}
\label{tab:performance_comparison}
\resizebox{\linewidth}{!}{%
\begin{tabular}{l|cccc|cccc}
\toprule
\multirow{2}{*}{Methods} & \multicolumn{4}{c|}{BRIGHT Dataset} & \multicolumn{4}{c}{Delta-SN6 Dataset} \\ \cline{2-9}
 & $F_1^{bcd}$ & $F_1^{clf}$ & OA & mIoU & $F_1^{bcd}$ & $F_1^{clf}$ & OA & mIoU \\ \midrule
STSF-Net(w/o. PGFFM) & 88.63 & 73.32 & 95.69 & 65.64 & 89.63 & 93.27 & 99.12 & 84.93 \\
STSF-Net(w. PGFFM with SAM) & 89.74 & 73.37 & 95.95 & 66.86 & 92.72 & 95.09 & 99.37 & 88.16 \\
STSF-Net(w. PGFFM with DINOv3) & 88.85 & 73.21 & 95.90 & 66.72 & 93.11 & 92.65 & 99.40 & 88.51 \\
\rowcolor{lightgray!20}
STSF-Net(w. PGFFM with SAM2) & 91.71 & 75.83 & 96.10 & 67.91 & 94.60 & 95.27 & 99.54 & 91.33 \\\bottomrule
\end{tabular}}
\end{table}

To evaluate the applicability of PGFFM across different architectures, we integrated it into three structurally distinct models: GSTM-SCD, DamageFormer, and SiamCRNN. As shown in Table \ref{pgffm_comparison}, PGFFM yields consistent improvements across all three models on both datasets, with average mIoU gains of approximately 1.15$\%$ on BRIGHT and 1.84$\%$ on Delta-SN6. These results demonstrate that PGFFM serves as a plug-and-play cross-modal fusion module that adaptively optimizes feature fusion by injecting semantic priors.

\begin{table*}[h]
\centering
\caption{Performance comparison of different methods with and without PGFFM on BRIGHT and Delta-SN6 datasets.}
\label{pgffm_comparison}
\resizebox{\linewidth}{!}{%
\begin{tabular}{cccc *{4}{S[table-format=2.2]} *{4}{S[table-format=2.2]}}
\toprule
\multirow{2}{*}{\textbf{Type}} &
\multirow{2}{*}{\textbf{Methods}} &
\multirow{2}{*}{\textbf{PGFFM}} &
\multicolumn{4}{c}{\textbf{BRIGHT}} &
\multicolumn{4}{c}{\textbf{Delta-SN6}} \\
\cmidrule(lr){4-7} \cmidrule(lr){8-11}
& & & {$\mathbf{F_1^{bcd}}$} & {$\mathbf{F_1^{clf}}$} & {$\mathbf{OA}$} & {$\mathbf{mIoU}$} & {$\mathbf{F_1^{bcd}}$} & {$\mathbf{F_1^{clf}}$} & {$\mathbf{OA}$} & {$\mathbf{mIoU}$} \\
\midrule
\multirow{2}{*}{CNN} & SiamCRNN & $\times$ & 88.77 & 68.71 & 95.42 & 63.37 & 89.44 & 92.93 & 99.11 & 83.82 \\
                      & SiamCRNN & $\checkmark$ & \textbf{89.26} & \textbf{72.01} & \textbf{95.76} & \textbf{65.57} & \textbf{90.67} & \textbf{94.46} & \textbf{99.25} & \textbf{86.79} \\
\midrule
\multirow{2}{*}{Transformer} & DamageFormer & $\times$ & \textbf{89.92} & 72.22 & \textbf{95.95} & 66.09 & 91.38 & 94.81 & 99.37 & 88.12 \\
                              & DamageFormer & $\checkmark$ & 89.46 & \textbf{72.71} & 95.92 & \textbf{66.49} & \textbf{93.09} & 95.41 & \textbf{99.41} & \textbf{89.36} \\
\midrule
\multirow{2}{*}{Mamba} & GSTM-SCD & $\times$ & 90.51 & 73.96 & 95.88 & 66.83 & 91.42 & 95.31 & 99.32 & 87.04 \\
                        & GSTM-SCD & $\checkmark$ & \textbf{90.88} & \textbf{76.27} & \textbf{95.93} & \textbf{67.69} & \textbf{92.33} & \textbf{96.04} & \textbf{99.57} & \textbf{88.36} \\
\bottomrule
\end{tabular}}
\vspace{0.2cm}
\small
\end{table*}

\subsection{Influence of the modality-common and modality-specific features in change information representation}
\begin{figure}[!h]
    \centering
    \includegraphics[width=0.95\linewidth]{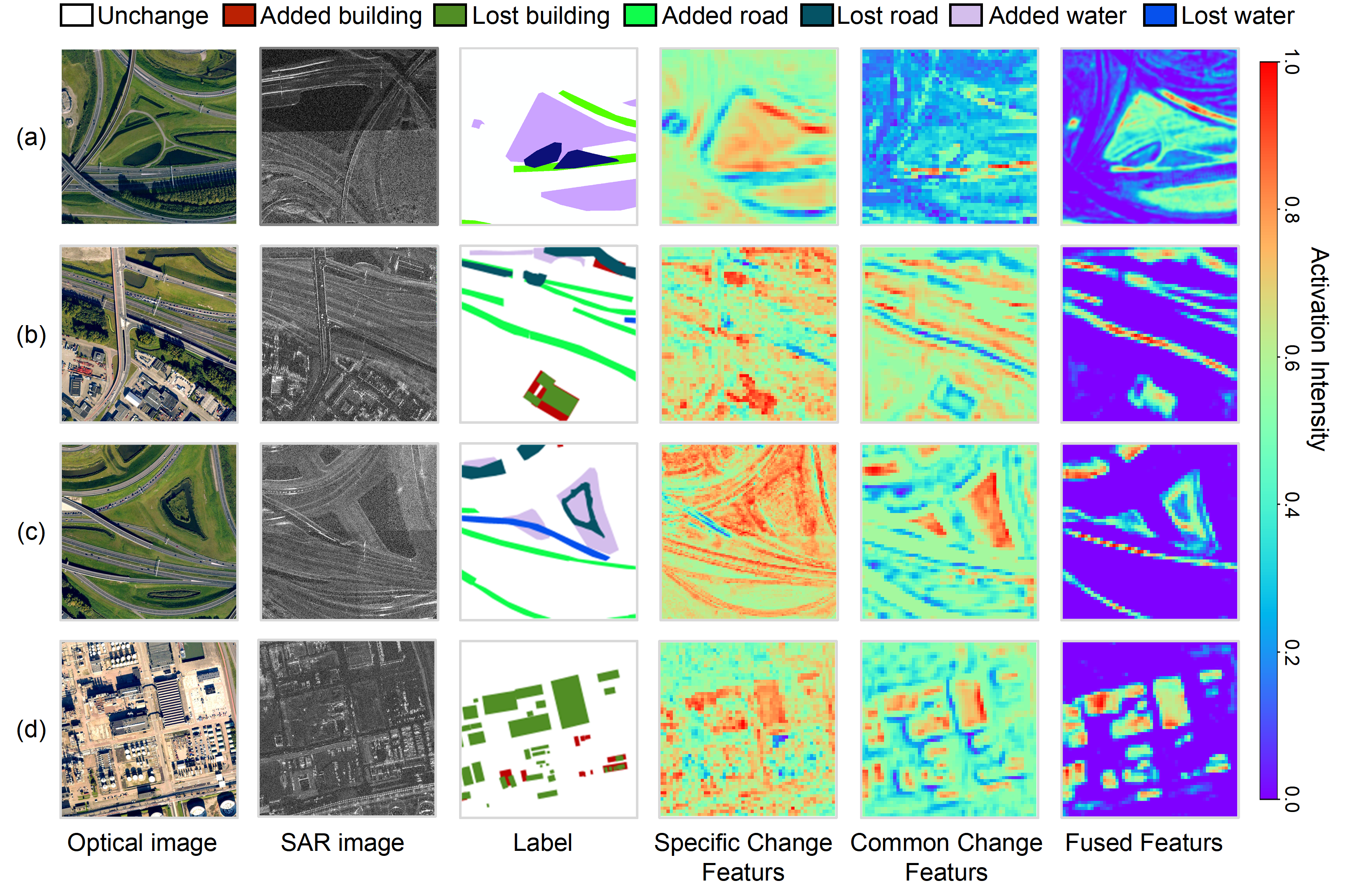}
    \caption{Visualization of specific features, common features, and the fused change feature on the Delta-SN6 dataset.}
    \label{fig12}
\end{figure}

To better understand how modality-common and modality-specific features influence change modeling, we visualize the corresponding change response maps. As shown in Fig.~\ref{fig12} (a) and (b), change responses derived from specific features exhibit strong activation in change regions (e.g., buildings and water bodies), but also produce noticeable noise due to modality discrepancies (e.g., road edges and bare soil). In contrast, Fig.~\ref{fig12} (c) and (d) illustrate that change responses based on common features are not active in unchanged regions yet show clear, coherent activations in changed regions. This property of low noise and high consistency is crucial for suppressing pseudo-changes. The fused change representation effectively integrates these complementary advantages, yielding high activation in true change regions and strong suppression in unchanged backgrounds, leading to a more complete and precise change representation.

\subsection{Analysis of model efficiency and complexity}
We compared the number of parameters (Params) and floating-point operations (FLOPs) across different CD methods. For a fair comparison, we used the \texttt{thop} package in PyTorch\footnote{\url{https://github.com/Lyken17/pytorch-OpCounter}} to estimate parameters and FLOPs under 3-channel 512×512-pixel inputs. As shown in Table \ref{tab:efficiency_comparison}, from the perspective of model complexity, STSF-Net has 63.37M trainable parameters and 191.42G FLOPs. While lightweight models such as M-UNet (14.49 M) and computationally efficient methods like ChangeOS (11.7 ms) exist, these approaches exhibit lower detection accuracy. STSF-Net incorporates a multi-scale feature enhancement and fusion strategy, achieving optimal detection accuracy while maintaining relatively efficient inference speed (51.60 ms). Its higher parameter count primarily stems from systematic modeling of multi-source, multi-scale remote sensing features, which is critical for accurately interpreting complex nonlinear relationships between optical and SAR imagery. Notably, STSF-Net remains significantly more computationally efficient than comparable methods (e.g., SiamAttnUNet requires 345.68 ms), demonstrating a favorable trade-off between computational resources and representational power.

\begin{table}[htbp]
\centering
\caption{Comparison of computational efficiency for different methods.}
\label{tab:efficiency_comparison}
\resizebox{0.8\linewidth}{!}{%
\begin{tabular}{l l c c c} 
\toprule
\textbf{Methods} & \textbf{Reference} & \textbf{Params (M)} & \textbf{FLOPs (G)} & \textbf{Inf. time (ms)} \\
\midrule
UNet & MICCAI'15 & 31.05 & 219.45 & 394.50 \\
DeepLabV3+ & ECCV'18 & 42.40 & 140.15 & 112.35 \\
SiamAttnUNet & ISPRS'21 & 60.95 & 494.76 & 345.68 \\
SiamCRNN & TGRS'20 & 54.37 & 135.02 & 118.78 \\
DamageFormer & IGARSS'22 & 48.16 & 202.50 & 216.15 \\
ChangeOS & RSE'21 & 50.60 & 117.53 & 11.7 \\
M-UNet & GRSL'22 &14.49   &70.95   &5.80\\
HRSICD & ISPRS'25 &20.05 &1814.23 &202.10 \\
Sigma & WACV'25 & 31.50 & 42.77 & 380.03 \\
GSTM-SCD & ISPRS'25 & 31.55 & 72.21 & 102.76 \\
FlowMamba & TSCVT'25 & 56.83 & 386.37 & 155.26 \\
\rowcolor{lightgray!30}
STSF-Net & \textit{Proposed} & 63.37 & 191.42 & 51.60 \\
\bottomrule
\end{tabular}}
\vspace{0.2cm}
\small
\end{table}

\section{Conclusion}
MMCD in RS leverages complementary observation data to enhance the timeliness of change analysis. However, significant challenges are posed in developing advanced MMCD methods, including the scarcity of dedicated MMCD benchmarks and a limited understanding of how modality-specific and modality-common features jointly represent changes. To address these issues, we propose an MMCD framework that explicitly disentangles and fuses the two types of features under the guidance of semantic priors from the pretrained foundation model SAM2. In greater detail, we first adopt a fine-tuned SAM2 together with SwinTransformer to capture rich modality-specific features. Then, spatio-temporal common features are exploited and enhanced via graph convolutional networks and attention mechanisms. Furthermore, we introduce Delta-SN6, the first publicly available multiclass MMCD dataset with sub-meter spatial resolution. It provides fully polarimetric SAR and optical images covering diverse land cover types, offering a comprehensive benchmark for multimodal change analysis.

The proposed method achieves the state-of-the-art accuracy across three MMCD benchmark datasets. Notably, on the Delta-SN6 dataset, it outperforms literature approaches by 3.18$\%$ in $F_1^{bcd}$ and 3.21$\%$ in mIoU. Ablation studies confirm the effectiveness of each module, especially the prior-guided fusion module. These results demonstrate that the joint modeling of modality-specific and modality-common features systematically improves MMCD accuracy and robustness. The underlying mechanism lies in the complementary and synergistic effect of decomposed modality features: modality-specific features preserve the unique physical details observed by each sensor that highlight subtle changes, while cross-modal common features capture consistent high-level semantics and suppress cross-modal noise. In future research, we plan to develop a unified lightweight MMCD framework via knowledge distillation and dynamic feature selection, enabling efficient CD across heterogeneous modality pairs.

\section*{Acknowledgements}
This work was supported by the Program of National Natural Science Foundation of China under Grant 42201443, and in part by the Henan Natural Science Foundation Youth Project under Grant No. 42001283.

\section*{Data Availability}
The BRIGHT dataset is openly available in the Zenodo repository at \href{https://doi.org/10.5281/zenodo.14619797}{https://doi.org/10.5281/zenodo.14619797} (Chen et al., 2025); the Wuhan dataset is openly available in the GitHub repository at \href{https://github.com/GeoZcx/A-Domain-Adaption-Neural-Network-for-Change-Detection-with-Heterogeneous-Optical-and-SAR-Remote-Sens}{https://github.com/GeoZcx/A-Domain-Adaption-Neural-Network-for-Change-Detection-with-Heterogeneous-Optical-and-SAR-Remote-Sens} (Zhang et al., 2022). The Delta-SN6 dataset and the source code of STSF-Net will be publicly available at \href{https://github.com/liuxuanguang/STSF-Net}{https://github.com/liuxuanguang/STSF-Net} upon publication, and are available from the first author (Xuanguang Liu) upon reasonable request.

\bibliography{reference}
\end{sloppypar}
\end{document}